\documentclass[floatsintext]{apa7}

\usepackage{lipsum}
\usepackage{amsmath}
\usepackage[american]{babel}
\usepackage{tabularx}
\usepackage{csquotes}
\usepackage{comment}
\usepackage{multirow}
\usepackage[style=apa,sortcites=true,sorting=nyt,backend=biber]{biblatex}
\DeclareLanguageMapping{american}{american-apa}
\title{Using novel data and ensemble models to improve automated labeling of Sustainable Development Goals}

\shorttitle{SDG labeling systems}

\author{Dirk U. Wulff$^{1, 2}$, Dominik S. Meier$^{1}$, and Rui Mata$^{1}$}
\affiliation{$^1$University of Basel\\$^2$Max Planck Institute for Human Development}

\leftheader{Wulff, Meier, Mata}


\authornote{
   \addORCIDlink{Dirk U. Wulff}{0000-0002-4008-8022}
   \addORCIDlink{Dominik S. Meier}{0000-0002-3999-1388}
   \addORCIDlink{Rui Mata}{0000-0002-1679-906X}

Correspondence concerning this article should be addressed to Dirk U. Wulff, Department of Psychology, University of Basel, Missionsstrasse 60-62, 4055 Basel, Switzerland.  E-mail: dirk.wulff@gmail.com}

\begin{document}
\maketitle



\textbf{A number of labeling systems based on text have been proposed to help monitor work on the United Nations (UN) Sustainable Development Goals (SDGs). Here, we present a systematic comparison of systems using a variety of text sources and show that systems differ considerably in their specificity (i.e., true-positive rate) and sensitivity (i.e., true-negative rate), have systematic biases (e.g., are more sensitive to specific SDGs relative to others), and are susceptible to the type and amount of text analyzed. We then show that an ensemble model that pools labeling systems alleviates some of these limitations, exceeding the labeling performance of all currently available systems. We conclude that researchers and policymakers should care about the choice of labeling system and that ensemble methods should be favored when drawing conclusions about the absolute and relative prevalence of work on the SDGs based on automated methods.}  

Monitoring work on the SDGs is crucial for their advancement and one promising approach is to screen the increasing amount of digitally available text using automated, natural language processing methods. This approach has taken hold, for example, in scientometric efforts that monitor the SDGs in academic publications \parencite[e.g.,][]{Jayabalasingham:gfBs-pcB,aurora, Bautista2019,duran_silva_nicolau_2019_3567769,sdsn}. One such system has identified millions of SDG-related academic publications and is already used by the Times Higher Education Impact Rankings to rank over 1,000 universities worldwide according to their SDG-related outputs. This example illustrates the important role that SDG labeling from text can play in academic research and funding policy \parencite{Smith.2021}. 

The increased and widespread use of SDG labeling systems should be accompanied by efforts to validate the different approaches and ensure that their predictions accurately reflect the SDGs. However, systematic and comprehensive evaluations of the accuracy of these labeling systems are largely lacking. Crucially, the few existing results indicate some striking differences in the predictions of the different labeling systems \parencite[e.g.,][]{armitage2020mapping, pukelis2020osdg,schmidt_felix_2021_4964606}. 

Three key reasons prevent rigorous evaluation of the different SDG labeling systems. First, most labeling systems have been developed for and tested with specific proprietary citation databases thus limiting their portability to other text sources. Second, there has been a paucity of publicly available data that can be used to validate the predictions of different approaches. Third, systems have often been analyzed in isolation without systematic comparisons involving various performance metrics or assessment of biases. 

In this paper, we aimed to help overcome these limitations by relying on our recently developed open-source R package, text2sdg \parencite[(\href{http://text2sdg.io/}{text2sdg.io})][]{meier2021text2sdg}, which can be used with any text source, and collating various labeled and non-labeled data sources to provide a comparison of seven existing SDG labeling systems using a comprehensive set of performance metrics. We provide a comparative evaluation of seven labeling systems; namely, the Aurora \parencite[]{aurora}, Elsevier \parencite[][]{Jayabalasingham:gfBs-pcB}, SIRIS \parencite[]{duran_silva_nicolau_2019_3567769}, Auckland \parencite[]{auckland}, SDGO \parencite[][]{Bautista2019}, and SDSN \parencite[Sustainable Development Solutions Network;][]{sdsn} systems implemented in the text2sdg R package and, additionally, OSDG.ai \parencite[]{osdgai}, a publicly available tool. These systems account for a majority of, albeit not all \parencite[cf.][]{fane_briony_2022_6951807}, systems currently available for automated detection of SDGs from text. Finally, we also aimed to assess the potential of using an ensemble approach that integrates several of the existing systems. Ensemble models can potentially improve accuracy and generalizability by considering the predictions of multiple models, each of which may have different biases, resulting in a more balanced and representative prediction. 

Our contribution is structured as follows. We first introduce and use three labeled data sets containing hand-coded labels to evaluate the categorizations of the seven automated labeling systems against those of human experts. The data sets cover different text sources, including titles and abstracts of academic publications, as well as news articles, thus significantly increasing the scope of sources considered in past work \parencite[e.g.,][]{armitage2020mapping, pukelis2020osdg}. Crucially, we use a number of different metrics to compare the different labeling systems (e.g., sensitivity, specificity, overall accuracy) thus providing a comprehensive assessment of the systems' strengths and weaknesses. Second, we compare the predictions of the SDG labeling systems across data sets to reveal SDG-specific biases; that is, whether different systems tend to make differential categorizations for specific SDGs. The potential for bias is particularly important concerning the application of automated labeling systems to make relative statements about the presence of specific SDGs. For example, the use of biased systems in the detection of some SDGs could lead to an incorrect assessment of investment in some domains (e.g., health) relative to others (e.g., education). Third, we introduce and leverage several novel unlabeled data sets (e.g., Disneyland reviews, cooking recipes, math lectures, random text) to better understand the labeling systems' susceptibility to the type and amount of text analyzed. Fourth, and finally, we explore whether ensemble-modeling approaches can address some of the potential limitations of existing SDG labeling systems. 

\section{Results}

In the following subsections, we first compare the seven labeling systems on a variety of metrics on three labeled data sets covering texts from academic publications and news articles. Second, we assess to what extent the different labeling systems show biases concerning different SDGs. Third, we assess the susceptibility of the different labeling systems to produce false positives as a function of the length of the text source in novel unlabeled data sets. Fourth, and finally, we assess the potential of ensemble models that integrate the different labeling systems to address potential limitations of individual labeling systems.

\subsection{SDG labeling systems differ in their sensitivity--specificity trade-offs}


Our first set of analyses consisted of comparing seven labeling systems to generate predicted labels for documents from three labeled data sets. All three data sets provide human judges' ratings of SDGs for different document types: the first data set consists of over 10,000 titles of academic publications \parencite[\textit{titles};][]{vanderfeesten_maurice_2020_3813230}, the second data set consists of over 30,000 abstracts of academic publications \parencite[\textit{abstracts}; ][]{osdg_2021_5550238}, and the third data set consists of over 9,000 news articles scraped, with permission, from the SDG knowledge hub website \parencite[\textit{news articles}; \url{https://sdg.iisd.org}][]{wulff_sdghub}. The data sets differ in a number of respects, most notably\textbf{} in the number of words per document, with titles, abstracts, and news articles being composed of 18, 90, and 673 words per document on average, respectively, and in the number SDGs they were evaluated for, with titles and abstracts having each been evaluated for only one of the SDGs and news articles having been evaluated for all 17 SDGs.

\begin{figure*}[h!]
    \centering
    \includegraphics[height=1\textheight, width=1\linewidth, keepaspectratio]{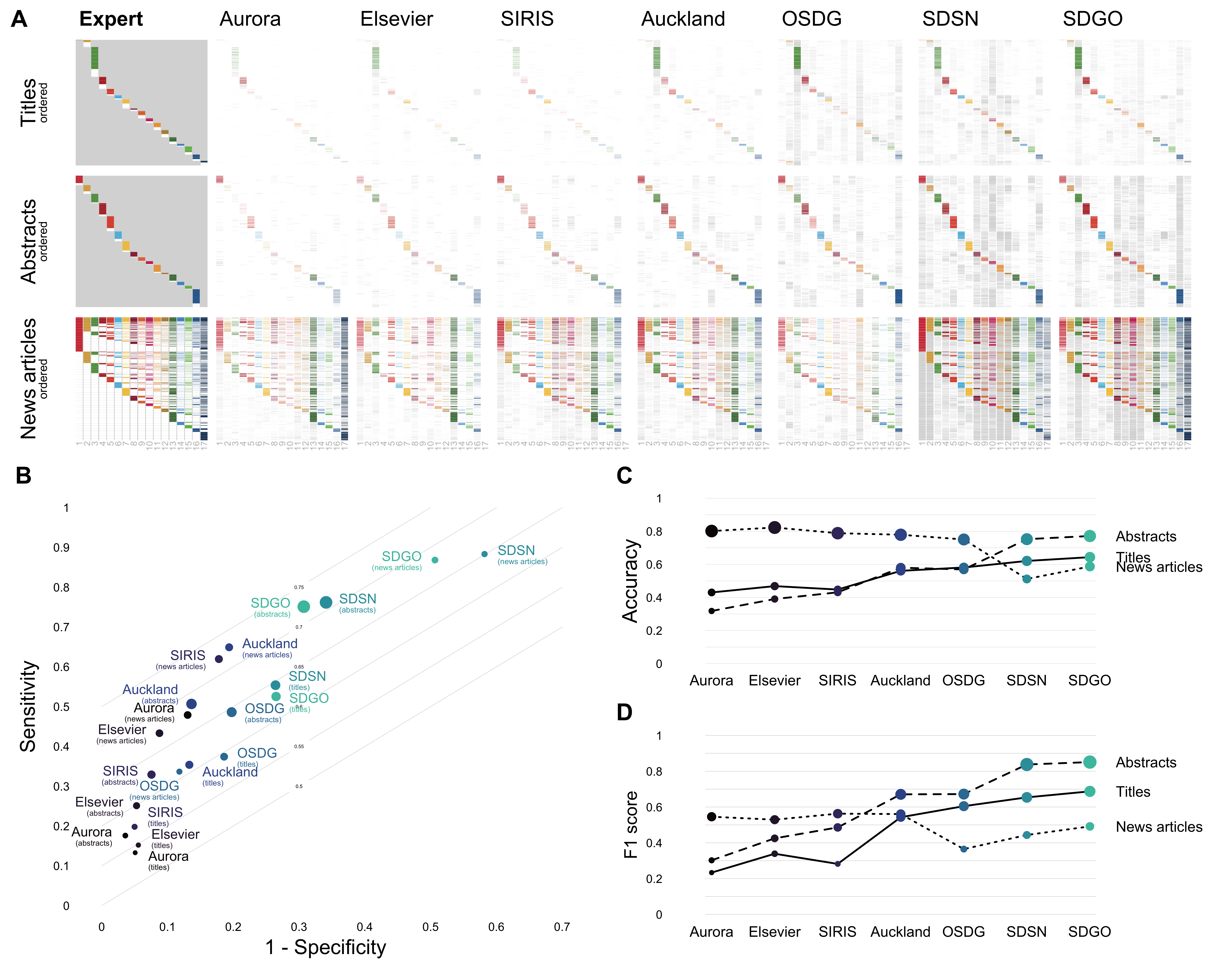}
    \caption{Performance of SDG labeling systems. Panel A shows the SDG labels assigned by experts and seven labeling systems for three data sets consisting of titles of research articles (Titles), abstracts of research articles (Abstracts), and online news articles (News Articles), respectively. The colored lines represent the assigned SDG by either experts or SDG labeling system and gray lines represent missing labels (for experts) or false negatives (for SDG labeling systems). Panel B shows the performance of the seven systems across the three data sets by plotting the systems' sensitivity against 1-specificity forming an ROC (receiver operating characteristic) plot. Lines in the background show several levels of balanced accuracy. Panel C and D show the performance of the seven systems across the three data sets in terms of accuracy and the F1 score.}
    \label{fig:performance}
\end{figure*}

Figure \ref{fig:performance} presents the main classification results separately for the three data sets (\textit{titles}, \textit{abstracts}, \textit{news articles}), with panel A showing a breakdown of SDG labels per document. To better assess the differences between each data set, it is helpful to look initially at the first column of each subpanel in Figure \ref{fig:performance}A, which contains information on the expert ratings for each data set. For the \textit{titles} data set, a large majority of documents were judged to contain one SDG (63\%). In the \textit{abstracts} data set, a majority of documents were assigned an SDG (80.2\%). In turn, in the \textit{news articles} data set, all of the documents (100\%) were assigned one or more SDGs. As a whole, across data sets, only a minority of documents have not been assigned any SDG by the experts, which, as we discuss below, can limit the validation of SDG labeling systems. 

The remaining columns in Figure \ref{fig:performance}A depict the labels assigned to each document by the labeling systems and, therefore, allow a first comparison between systems and the human experts. Visual inspection suggests that all systems showed reasonable accuracy in terms of recovering the "true" SDG labels assigned by experts. This is illustrated by the similar pattern of colored stripes across experts and the seven labeling systems. Nevertheless, the systems showed considerable differences in their ability to detect SDGs, as illustrated by the relative intensity of the colored stripes and gray stripes, respectively. The results below quantify these differences using common performance metrics. 


We compared the seven labeling systems quantitatively using a number of common metrics typically used in categorization problems (i.e., sensitivity, specificity, accuracy, F1 score). Sensitivity measures a system's ability to correctly identify true SDG text sources as positive whereas specificity measures a system's ability to identify non-SDG sources as negative. Ideally, SDG labeling systems should be both sensitive and specific but there is often a trade-off between the two. Considering both metrics can help compare the systems and their criteria for dealing with trade-offs between the two. In practice, we do so by visually comparing such metrics in an ROC (receiver operating characteristic) space as well as considering general measures of categorization performance, such as overall accuracy and a composite measure of sensitivity and specificity (i.e., F1 score) in the different data sets. 

As can be seen in Figure \ref{fig:performance}B, the systems differ substantially in their trade-off between sensitivity and specificity or, in other words, in how conservatively they assign SDGs. This is shown by the fact that the systems' performance---in terms of sensitivity and specificity---varies mostly along the diagonal of the ROC plot. We also observe differences in overall accuracy but these are not stable across data sets (Figure \ref{fig:performance}C), which is due to the different levels of conservatism of the different labeling systems: More conservative systems (e.g., Elsevier) outperform liberal systems for news articles, in which many labels are negative, and vice versa for abstracts and titles, where almost all labels are positive. These patterns hold for the alternative measure of performance, the F1 score (Figure \ref{fig:performance}D). 

All in all, these results point to labeling systems being differently conservative; that is, they solve the sensitivity--specificity trade-off differently. As a consequence, labeling systems do best in different data sets and it is difficult to identify a single best-performing model. The results further point to a limitation of currently available data sets that include only a small proportion of non-SDG-related documents, which introduces difficulties in assessing systems' susceptibility to producing false positives. We return to this point below when we introduce novel synthetic data sets.  

\section{Biases in SDG labeling systems distort SDG profiles}

Biases in SDG labeling systems can lead to inaccurate representation of the prevalence and importance of different SDGs. If a system is more sensitive to certain SDGs than others, it will overestimate the prevalence of those SDGs and underestimate the prevalence of the others. This results in a misleading picture of the work being done to address the SDGs. Biased systems may also create an unfair advantage or disadvantage for certain organizations or groups. For example, if a method is more sensitive to certain SDGs than others, organizations may aim to portray themselves as focusing on those SDGs by relying on such systems. Finally, biased systems may create confusion or mistrust among stakeholders. If different methods produce significantly different results, it may be difficult for stakeholders to know which results to trust. This can lead to confusion or mistrust in the results, which could ultimately undermine the credibility of the work being done to address the SDGs.

We estimated SDG-specific biases by comparing the relative frequency of SDGs between predicted and observed labels. Specifically, we calculate $bias = \frac{predicted-observed}{observed}$. We then evaluate the profile bias of a given system by correlating the biases across data sets. We also compare the SDG profiles obtained from the seven labeling systems to the SDG profile of experts. This analysis amounts to assessing the similarity between the profile across SDGs identified by experts and the profile identified by each of the labeling systems. 

\begin{figure*}[!]
    \centering
    \includegraphics[height=1\textheight, width=1\linewidth, keepaspectratio]{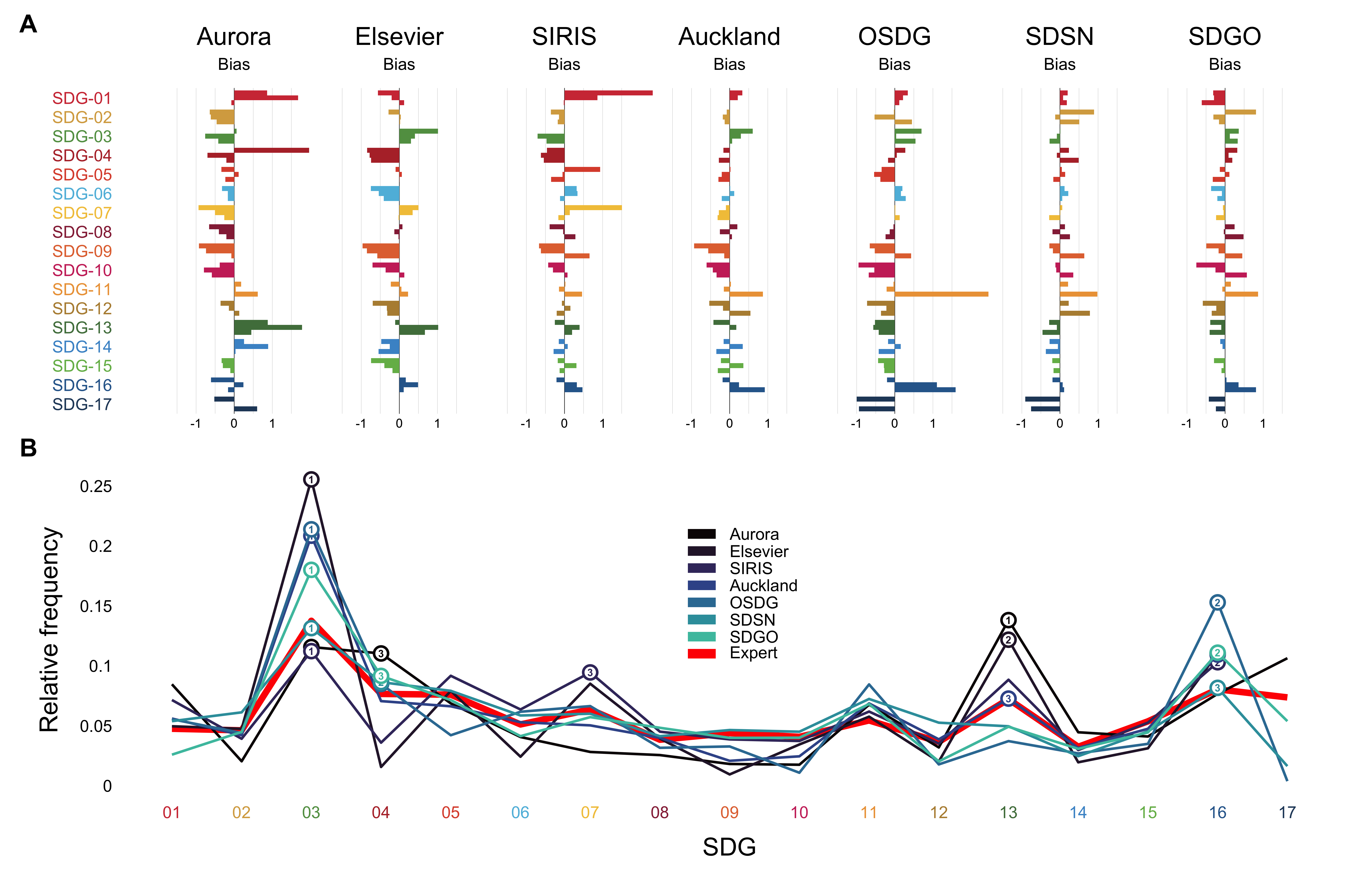}
    \caption{Biases in SDG classifications. Panel A illustrates the biases of SDG labeling systems in classifying each of the 17 SDGs. Biases are calculated as the difference between the observed and predicted SDG proportion for a given data set divided by the observed proportion. A positive bias means that a given SDG was assigned more often than its relative frequency in expert judgments and vice versa for a negative bias. The three vertical stripes for each SDG and system correspond to the titles, abstracts, and news articles data sets. Panel B shows the profile of expert (red) and system labels generated by averaging the relative frequencies of SDGs across data sets. The circles and numbers highlight the three most prevalent SDGs per system. }
    \label{fig:bias}
\end{figure*}

Figure \ref{fig:bias}A shows the biases of the different labeling systems. Visual inspection suggests that some biases appear systematically across data sets. For example, considering the Aurora labeling system, the pattern suggests consistent underestimation of SDGs 2, 6, 7, 8, 9, and 10 across data sets but overestimation of SDG 13. For Elsevier, one observes consistent underestimation of SDGs 4, 6, 9, 12, 14, and 15 but overestimation of SDGs 3 and 16. To quantify the systems' profile biases we correlated the SDG biases between titles and news articles and between abstracts and news articles. We did not consider the correlation between titles and abstracts because of a moderate correlation in expert profiles for these two sources. The strongest average profile bias was observed for Elsevier ($\bar{r}=0.72$), followed by OSDG ($\bar{r}=0.41$), Aurora ($\bar{r}=0.38$), SDGO ($\bar{r}=0.31$), SDSN ($\bar{r}=0.2$), Auckland ($\bar{r}=0.08$), and SIRIS ($\bar{r}=-.05$). 

Figure \ref{fig:bias}B shows that these biases imply substantial differences in SDG profiles derived from the different labeling systems relative to the one derived from experts. This difference is most pronounced for SDG 3 (Good health and well-being). A number of systems overestimate the relative frequency of SDG 3, with Elsevier giving SDG 3 twice as much weight ($.26$) as experts ($.13$). In turn, SDGs 9 (Industry, Innovation, and Infrastructure) and 10 (Reduced inequalities) tend to receive less weight than assigned by experts. To further evaluate profile fidelity, we calculated the Spearman's rank correlation coefficient between expert and system relative frequencies separately for the three data sets. We found Auckland to have the highest average correlation ($\bar{r} = .76$), followed by SDGO ($\bar{r}=.72$), Elsevier ($\bar{r}=.70$), SIRIS ($\bar{r}=.67$), Aurora ($\bar{r}=.57$), and, finally, OSDG ($\bar{r}=.54$). 

To summarize, we find that systems appear to have different biases, with a number of these underestimating or overestimating the presence of specific SDGs relative to experts. These biases can be important for the relative assessment of work on the SDGs and emphasize that the labeling systems should not be used interchangeably. 

\subsection{SDG labeling systems can produce many false positives when applied to large text sources}

Existing validation data sets---such as those used above to test the accuracy of SDG labeling systems---may not accurately reflect the performance of SDG labeling systems in many real-world applications that involve larger samples of text. As mentioned above, existing data sets only include a small proportion of documents unrelated to the SDGs, which can lead to an overestimation of the accuracy of the labeling systems relative to applications in which labeling systems are tested on a diverse range of documents that can potentially contain many false positives. To address this issue, we conducted evaluations of the labeling systems using data sets that are ostensibly unrelated to SDGs in order to better understand the systems' tendency to produce false positives; in particular, as a function of text length.

In this analysis, we used existing and synthetic data sets; specifically, three natural language data sets consisting of Disneyland reviews ($N = 42,656$), cooking recipes ($N = 82,245$), and math lectures ($N = 860$). In addition, we generated synthetic texts by sampling from a word frequency list derived from Wikipedia and generating documents containing 10, 100, 1,000, and 10,000 randomly sampled words.

\begin{figure*}[tb!]
    \centering
    \includegraphics[height=1\textheight, width=1\linewidth, keepaspectratio]{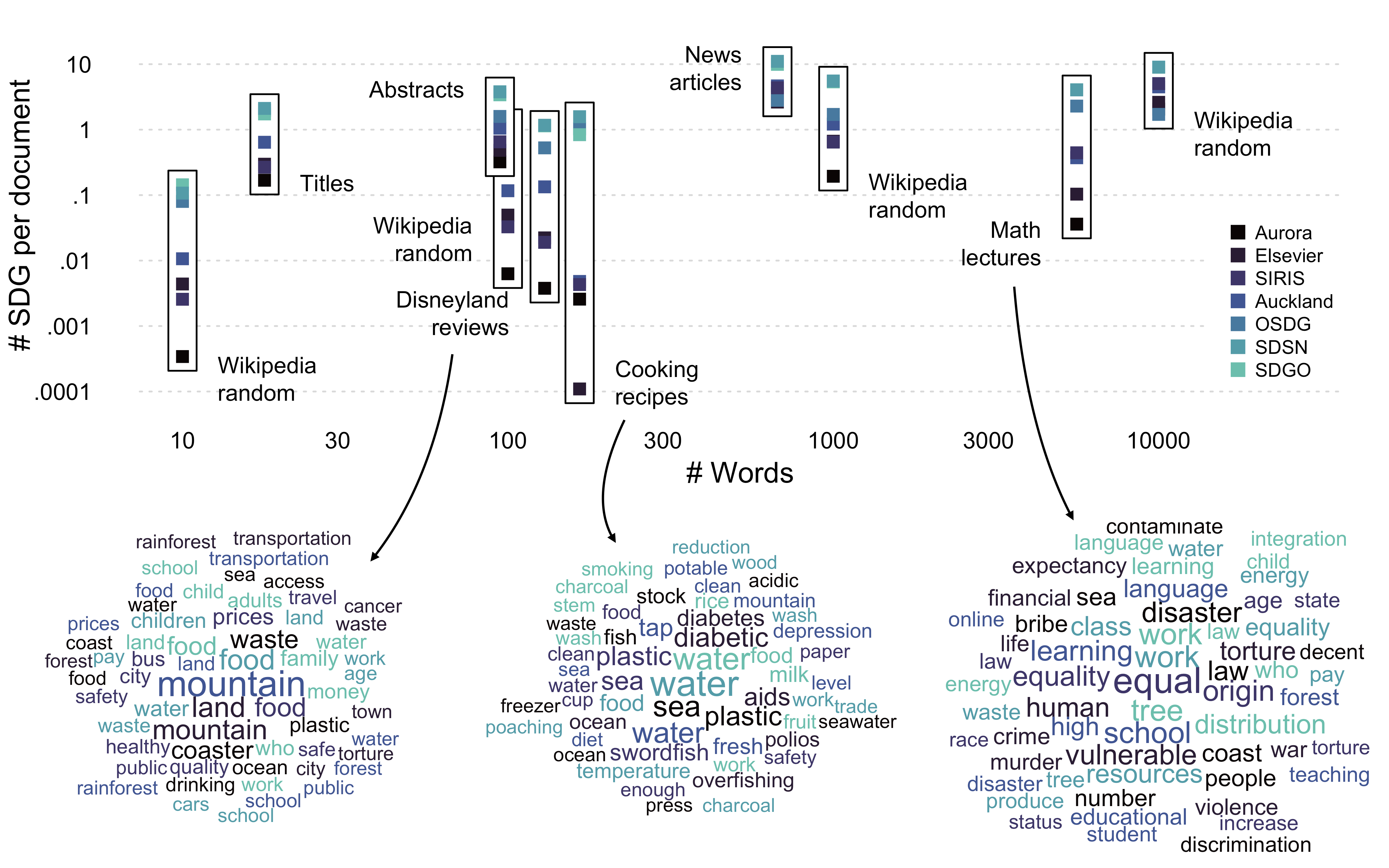}
    \caption{False positive predictions of SDG labeling systems. The figure shows the number of SDGs assigned by the seven systems to three natural language data sets unrelated to SDGs (Disneyland reviews, cooking recipes, and math lectures), four synthetic data sets of different lengths created from Wikipedia word frequencies (Wikipedia random), and three expert-labeled data sets (Titles, Abstracts, News Articles), as a function of the number of words per document. The word clouds at the bottom consists of the keywords that triggered the assignment of SDGs for the three natural language data sets, with size coding the frequency of keyword hits and the color coding the SDG labeling system.}
    \label{fig:fas}
\end{figure*}

Figure \ref{fig:fas} presents the results of the SDG labeling systems for the various data sets. The plot shows the number of SDGs identified in each document by each system, plotted against the average length of the documents in the respective data set. The plot includes data from the novel data sets (those likely not related to the SDGs), as well as the three expert-labeled data sets. We use the number of SDGs per document as a proxy for the false-positive rate. Consequently, by comparing the number of SDGs identified by the systems across different data sets and document lengths, we hope to understand the systems' susceptibility to producing false positives and identify any patterns or trends in their accuracy as a function of document length. 

Several noteworthy results emerge. First, all systems produce false positives for all types of data sets. Second, the tendency of systems to produce false positives is in line with their level of conservatism, which we discussed above. For example, Aurora and Elsevier appear to be very conservative in identifying SDGs, leading to an overall low false-positive rate. Third, across all systems the tendency to detect SDGs increases considerably when the length of the texts increases. Specifically, whereas systems produced between $.0003$ (Aurora) and $.144$ (SDGO) for synthetic texts of length 10, systems produced between $1.71$ (OSDG) and $9.03$ (SDSN) SDGs per document for synthetic texts of length 10,000. 

The word clouds illustrate the keywords that triggered the assignment of SDGs for the three natural language data sets (Disneyland reviews, cooking recipes, math lectures). As can be seen, the keywords relate to many frequent topics that fit the respective SDGs but also the mundane contexts of the three data sets. These word clouds can be seen to highlight the limitations of query-based approaches that do not control for the frequency of keywords in natural language.     

\subsection{Trained ensemble models alleviate the shortcomings of existing labeling systems}

Previous sections illustrated serious shortcomings of existing SDG labeling systems. Labeling systems differ more in conservatism than in accuracy, they have considerable SDG-specific biases, and they commit many false positives when the length of documents increases. However, these shortcomings did not affect systems uniformly---some systems were more prone to bias whereas others were more prone to false alarms. We leverage this fact to train ensemble models of SDG labeling systems. 

We consider the six publicly available labeling systems implemented in text2sdg and document length as features and combine them in a random forest model. We train the model on the three expert-labeled data sets and, to control the false-positive rate of the model, on synthetic data sets generated from Wikipedia word frequency, matching in length the documents of the three labeled data sets. We train the model by assigning equal weight to the three labeled data sets and vary the weight of the synthetic data to vary the focus on reducing false positives. 

\begin{figure*}[t!]
    \centering
    \includegraphics[height=1\textheight, width=1\linewidth, keepaspectratio]{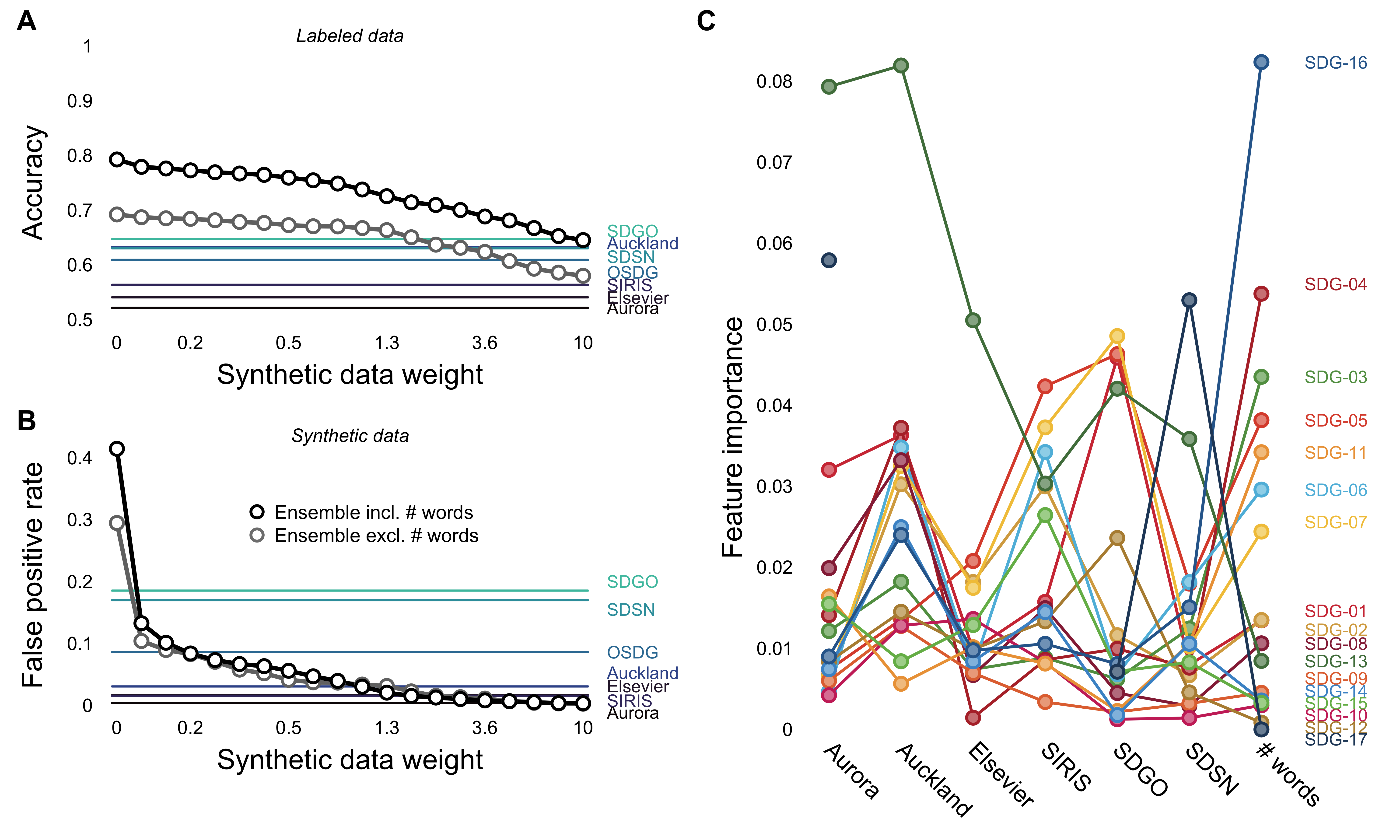}
    \caption{Ensemble model performance. Panel A and B illustrate the performance of ensemble models drawing on the classifications of the six labeling systems relative to the labeling systems on their own. Panel A shows the accuracy for the three expert-labeled data sets as a function of the amount of training weight given to synthetic non-SDG documents generated from Wikipedia word frequencies. Panel B shows the false-positive rate as a function of the training weight. In both panels, the black line shows the performance of an ensemble model that includes document length as a feature, whereas the gray line corresponds to a model that does not include document length. Panel C shows the SDG-specific feature importance for the ensemble model that includes document length and has been trained with a non-SDG data weight of 1.}
    \label{fig:ensemble}
\end{figure*}

Figure \ref{fig:ensemble} shows the performance of two ensemble models including (black) and excluding (gray) document length as a predictor compared to the six labeling systems used to train the ensemble model and the OSDG labeling system. We find that for a wide band of moderate synthetic data weights the ensemble model achieves a higher average out-of-sample accuracy than all individual labeling systems, while committing only as many false alarms as the most conservative labeling systems. Furthermore, we find that including document length as a predictor substantially improves the performance of the ensemble model. 

We analyzed whether the ensemble model suffers from the same sensitivity--specificity trade-off and found that this was not the case. Specifically, for a synthetic data weight of $1$, we find that the ensemble model is on a par with the best-performing conservative model for the news article data (Ensemble: $Accuracy = .83$, $F1 = .54$; Elsevier: $Accuracy .82$, $F1 = .46$) and also with the best-performing liberal model for the titles and abstract data (Ensemble: $Accuracy = .69$, $F1 = .71$; SDSN: $Accuracy = .69$, $F1 = .73$), implying an even performance across data sets. We also evaluated the profile bias and fidelity of the ensemble model. We observed a profile bias at the lower end of the individual systems ($\bar{r} = .14$) and a expert profile fidelity far outperforming the individual systems ($\bar{r} = .92$).  

The ensemble model can further be used to understand the usefulness of the individual labeling systems through an analysis of feature importance. Figure 4C shows the permutation-based feature importance separately for the different SDGs. It can be seen that importance varies considerably across SDGs, which highlights not only differences in the quality of labeling systems across the SDGs but also the lack of comparable data on the different SDGs. Despite the considerable variance across SDGs, the ensemble model preferred to rely on some systems more than others. Specifically, Auckland, SDGO, and SDSN received higher feature importance relative to the Aurora, Elsevier, and SIRIS systems. 

\section{Discussion}


We aimed to compare a number of existing automated labeling systems that have been proposed to identify work on the Sustainable Development Goals (SDGs) from text. Systematic comparison of these systems, including their relative performance and potential biases, is a crucial step towards identifying the most reliable and accurate tools and, furthermore, building confidence in the use of automated methods for monitoring work on the SDGs.

We compared seven systems using a variety of text sources---including research papers, news articles, and non-SDG-related texts---and a variety of metrics. Our results suggest that the existing labeling systems differ in accuracy and that their performance varies considerably across text sources. These differences are due to the systems' differences in their specificity (true-positive rate) and sensitivity (true negative rate). Additionally, the systems have different biases that can have an impact on the overall profile of the SDGs identified, with some systems emphasizing specific SDGs (e.g., health) relative to experts. Our finding of biases is important because it reveals a potential for misleading representation of work on the SDGs that could create confusion about the relative investment in different SDGs and even undermine trust in the use of automated methods. Some researchers have pointed out how institutional rankings using several, often non-transparent criteria can lead to a lack of convergent validity between rankings and associated confusion \parencite[]{berg_aggregate_2022}. The increasing reliance on automated systems for ranking institutions' contributions to the SDGs requires reliable systems that aim to reduce bias. Finally, and more broadly, our results suggest that labeling systems should not be used interchangeably. 

One alternative to the use of single systems is the use of ensemble models that pool multiple labeling systems. Our results suggest that ensemble models can overcome some of the limitations of individual labeling systems. In particular, our results suggest that an ensemble approach is able to achieve higher performance compared to existing systems, and is less susceptible to biases and variations in the type and amount of text analyzed. All in all, these results suggest that ensemble models may be a good alternative to existing systems to detect SDGs in an automated fashion. Our ensemble model is freely available in our \href{https://www.text2sdg.io/}{text2sdg} R package. 

There are several limitations and opportunities for future research in the field of automated SDG labeling. First, none of the expert-labeled data sets used in our work allow for a good estimation of false positives, as these documents were not randomly or representatively sampled from the respective class of documents. We alleviated this problem by relying on synthetic data sets that were unrelated to SDGs but this approach does not account for the word co-occurrence patterns of actual negative SDG texts. Future work should focus on producing better validation data sets, ideally also from other domains beyond academic papers or news articles, if these methods are to be used beyond these domains, for example, concerning policy documents. 

Second, although we compared seven labeling systems that represent the majority of the existing systems for automated detection of SDGs from text, there are additional systems that we could not consider. Prominent among these are proprietary systems from Dimension.ai \parencite{fane_briony_2022_6951807} and novel versions from Elsevier. Overall, we believe that making such automated tools publicly and easily available would be an important step towards assessing and improving automated labeling of SDGs. 

Third, although the ensemble approach used in this study was effective in addressing the shortcomings of existing systems, it is still limited by the keywords used from the seven included systems. These keywords were selected using procedures specific to certain data sets, and it is likely that the current keyword set is incomplete for some or all of the SDGs. Once larger, more diverse, and representative expert-labeled data sets become available, it would be beneficial to learn relevant keywords from these expert labels using natural language processing techniques. One particularly promising approach is to fine-tune large-scale language models, which could not only provide high accuracy but also allow for the identification of effective keyword sets using explainable machine learning methods. Large language models can analyze the content and context of text to identify relevant keywords and phrases that are indicative of the SDGs, and can also incorporate information about the structure and organization of text, leading to a better understanding of the relationships between different concepts. Crucially, such models can be trained and fine-tuned in an iterative fashion to specific domains or languages, which can help to improve their performance and accuracy in specific contexts. Although such attempts exist, these have been built based on queries rather than the expert-labeled data \parencite{vanderfeesten_maurice_2022_5939866} thus developing well-labeled data sources remains a priority for moving such efforts forward.

In conclusion, we presented a comparison of existing labeling systems to identify work on the Sustainable Development Goals (SDGs) in text sources. Our approach was based on the use of novel data sources and several performance metrics. We found that current systems suffer from several shortcomings but that ensemble modeling techniques allow us to overcome some of the limitations of existing labeling systems. We demonstrate that an ensemble approach is able to achieve higher specificity and sensitivity compared to existing systems, and is less susceptible to biases and variations in the type and amount of text analyzed. Our findings have important implications for researchers and policymakers seeking to accurately monitor progress on the SDGs, and we recommend the use of ensemble approaches as best practice when drawing conclusions about the absolute and relative prevalence of work on the SDGs based on automated methods.

\subsection{Methods}


\subsubsection{SDG labeling systems}
Six of the systems analyzed are based on Lucene-style queries and vary considerably in complexity. The SDSN and SDGO systems are least complex because they only make use of OR-operations, implying that they assign an SDG as soon as a single keyword is matched. The SIRIS and, in particular, the Elsevier and Auckland systems are more complex as they additionally include AND-operations, meaning that multiple keywords must be present to trigger a match. The Aurora system is most complex because it further includes NEAR-operations, meaning that keywords must co-occur within a maximum distance to result in match. The seventh system developed by OSDG.ai \parencite[]{osdgai} is based on a machine learning model trained on the OSDG Community Dataset \parencite{osdg_2021_5550238}. 

\subsubsection{Labeled data}

The Aurora data set \parencite{vanderfeesten_maurice_2020_3813230} was created to validate the Aurora classification system. The part of the data we use consists of a survey where people had to indicate whether a research paper was relevant for a given SDG. Each of the 244 respondents did this for 100 papers randomly selected from a pool of research papers detected by the Aurora system as SDG relevant. 

The OSDG Community Dataset \parencite{osdg_2021_5550238} contains tens of thousands of text excerpts which were labeled by Community volunteers. To make this labeling more efficient, the volunteers only had to indicate whether or not a suggested label suited the text excerpt. Thus, the volunteers simply had to accept or reject a given SDG but were not asked to select one or more SDGs that might relate to the given text excerpt. Each text was rated by multiple volunteers 

The SDG Knowledge Hub data \parencite{wulff_sdghub} consists of news articles posted on the SDG Knowledge Hub website (sdg.iisd.org). This website was launched in October 2016 and is managed by the International Institute for Sustainable Development (IISD). It hosts news and commentary regarding the implementation of the SDGs. The news articles contain labels that show which SDGs they cover. These labels are assigned by the subject experts who write these news articles and confirmed by SDG Knowledge Hub editors. We downloaded 9,172 news articles that have been published on the website along with the assigned SDG labels. 

\subsubsection{Unlabeled data}

 The Disneyland reviews, cooking recipes, and math lectures were obtained from Kaggle. The Disneyland data set contains 42,656 reviews of Disneyland locations in Paris, California, or Hong Kong posted by visitors on Trip Advisor (see \href{https://www.kaggle.com/data sets/arushchillar/disneyland-reviews}{kaggle.com}). The cooking recipe data set contains 82,245 recipes scraped from food-related websites, such as \href{www.skinnytaste.com}{skinnytaste.com} (see \href{https://www.kaggle.com/data sets/extralime/math-lectures}{kaggle.com}). The math lecture data set contains 860 lectures posted on YouTube by institutions or creators covering 11 subjects, ranging from algebra to natural language processing, that are related to computer science and mathematics (see \href{https://www.kaggle.com/data sets/snehallokesh31096/recipe}{kaggle.com}). 

 The synthetic data sets were generated by concatenating words sampled at random based on the words' frequencies in the  (\href{https://en.wiktionary.org/wiki/Wiktionary:Frequency_lists}{Wikipedia corpus}. The synthetic texts, thereby, reflect the natural word frequency distribution found in natural language.  

\subsubsection{text2sdg}

To detect SDGs in these different texts, we used the text2sdg R package \parencite[\href{http://text2sdg.io}{text2sdg.io}][]{meier2021text2sdg}. text2sdg provides a common framework for implementing the different systems to detect SDGs in text and makes it easy to quantitatively compare and visualize their results. The text2sdg also makes available the ensemble model presented in this article. 
 
\subsubsection{Ensemble modeling}

 We recruited two types of algorithms, random forest \parencite[R package ranger][]{wright2015ranger} and extreme gradient boosting \parencite[R package xgboost]{chen2015xgboost}, to train ensembles of SDG labeling systems. We trained the algorithms separately for each SDG to predict the presence or absence of the SDG based on the predictions of six different SDG labeling systems implemented in \href{http://text2sdg.io}{text2sdg.io} and the number of words in the documents. Training and evaluation were performed using a repeated k-fold cross validation procedure. The models were trained using all three expert-labeled and synthetic data sets, with the latter matching the former in numbers and word lengths. The cases in the data sets were initially weighted by $1/N$ with $N$ being the number of cases in a data set to give each data set equal weight. Furthermore, we multiplied the weight of the synthetic data sets by a factor $k \in [0, 10]$ to vary the weight of synthetic data relative to the expert-labeled data. A factor of $k = 0$ means that the synthetic data receives no weight whereas a factor of $k = 10$ means that the synthetic data receive ten times as much weight as the expert-labeled data. Overall, we found the random forest to perform slightly better than the gradient-boosting algorithm; hence, we report the results of the random forest in the main text.      


\printbibliography

\appendix

\subsection{Data availability} 

The text2sdg R package is available on the Comprehensive R Network (\url{https://CRAN.R-project.org/package=text2sdg}). All data sets used in this analysis are publicly available. Links are included in the main text. 

\subsection{Acknowledgements}

 We are grateful to Laura Wiles for editing the manuscript. This work was supported by a grant from the Swiss Science Foundation (100015\_197315) to Dirk U. Wulff. 

\end{document}